\pdfoutput=1
\documentclass[11pt]{article}

\usepackage{ACL2023}

\usepackage{times}
\usepackage{latexsym}
\usepackage{csquotes}
\usepackage{booktabs}  
\usepackage[T1]{fontenc}
\usepackage[utf8]{inputenc}

\usepackage{microtype}
\usepackage{graphicx}
\usepackage{inconsolata}
\usepackage{verbatim} 
\usepackage{enumitem}
\usepackage[normalem]{ulem}
\usepackage{array}

\usepackage{xcolor}

\title{Tackling Hallucinations in Neural Chart Summarization}

\author{Saad Obaid ul Islam\textsuperscript{1,2,3}\quad Iza Škrjanec\textsuperscript{2}\quad Ondřej Dušek\textsuperscript{1}\quad Vera Demberg\textsuperscript{2} \\
  \textsuperscript{1}Charles University, Faculty of Mathematics and Physics, Prague, Czechia \\
  \textsuperscript{2}Saarland University, Saarbrücken, Germany \\
  \textsuperscript{3}Julius-Maximilians-Universität Würzburg, Germany \\
  \texttt{saadob12@gmail.com}, \texttt{\{skrjanec,vera\}@coli.uni-saarland.de}, \texttt{odusek@ufal.mff.cuni.cz}
  }

\makeatletter
\patchcmd{\ps@plain}{\thepage}{\textcolor{red}{\bf \thepage\  XXX\ remove me}}{}{}
\makeatother

\begin{document}
\maketitle
\begin{abstract}
Hallucinations in text generation occur when the system produces text that is not grounded in the input. In this work, we tackle the problem of hallucinations in neural chart summarization. Our analysis shows that the target side of chart summarization training datasets often contains additional information, leading to hallucinations.  We propose a natural language inference (NLI) based method to preprocess the training data and show through human evaluation that our method significantly reduces hallucinations. We also found that shortening long-distance dependencies in the input sequence and adding chart-related information like title and legends improves the overall performance. % of the system by 15.4 and 2.8 BLEU points on two chart summarization datasets.
\end{abstract}

\section{Introduction}
The task of generating a summary to accompany a chart is an instance of data-to-text generation and has a long tradition in natural language generation (NLG) \cite{Elzer-abrowser, ferres-igraphlite, demir-etal-2012-summarizing}. Recent neural models for chart summarization \citep{obeid2020chart,hsu2021scicap, zhu2021autochart, kanthara2022chart} carry the promise to be trainable from data and hence more versatile than approaches using manually constructed templates, and to produce more fluent text than previous statistical NLG systems. However, texts generated by state-of-the-art neural systems frequently include information which is not grounded in the input (``extrinsic hallucination''), or is even contradictory to it (``intrinsic hallucination''), see an example in Table \ref{tab:intro-example}. 
\begin{table}[]
    \small
        \includegraphics[width=\linewidth]{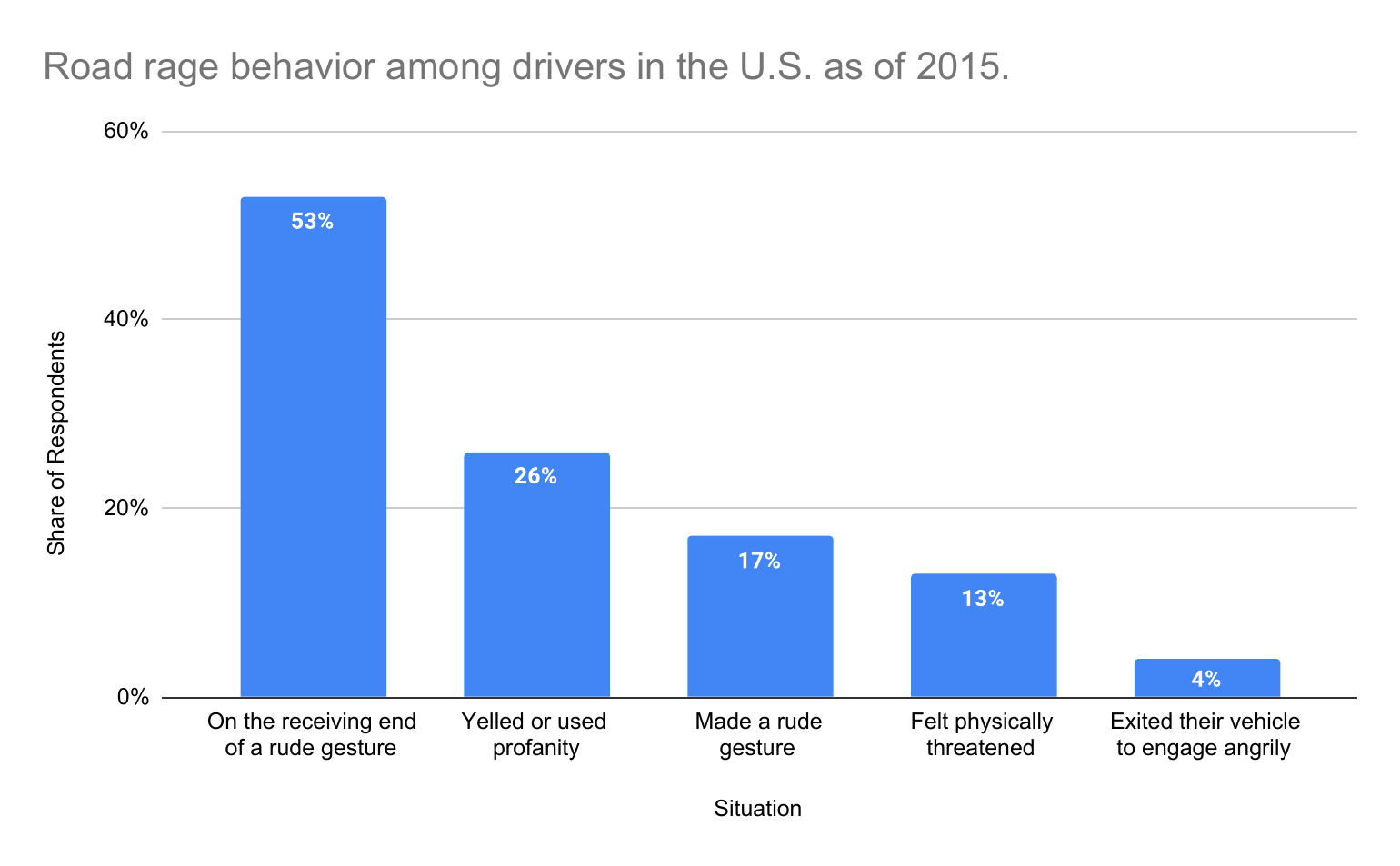}\\
        This statistic shows the road rage behavior of drivers in the United States as of 2015. \uline{Four percent of the drivers said they have been on the receiving end of a rude gesture.} \textit{The survey was conducted online and all the participants had a valid U.S. driving license.} 
    \caption{This output example from the chart-to-text NLG system by \citet{kanthara2022chart} includes \underline{intrinsic}, and \textit{extrinsic} hallucinations.}
    \label{tab:intro-example}
\end{table}

Hallucinations  in NLG \citep{koehn2017six,raunak2021curious} have been a concern in neural models for various tasks \citep{huang2021factual, lee2019hallucinations, rebuffel2022controlling}.
%In this work, we analyze the reasons for hallucinations in chart summarization and devise methods that help in reducing them.
We identify two reasons for hallucinations in chart summarization: (1) complexity and missing information in the input format of chart data; (2) presence of ungrounded information in chart summaries of the training data. Our contributions are as follows:
\begin{itemize}
    %\item We identify two reasons for hallucinations in chart summarization: (1) complexity and missing information in the input format of chart data; (2) presence of ungrounded information in chart summaries of the training data.
    \item We demonstrate the importance of providing more context and reducing long-distance dependencies in the linearized input format.
    \item We propose an NLI cleaning step to remove ungrounded information in the training data.
\end{itemize}
Our experimental code and model output will be released on Github under an open license.\footnote{\url{https://github.com/WorldHellow/Hallucinations-C2T}}
%\OD{This needs to be deanonymized}}

\section{Background and Related Work}

\subsection{Recent work in Chart Summarization}\label{chart-summarization}
Several chart summarization datasets and models were developed recently. % and made available to the public. 
\citet{obeid2020chart} created the Chart-to-Text data with English charts from statista.com (dubbed c2t-small in this paper). They model chart summarization as a data-to-text problem and adapt a transformer by \citet{gong-etal-2019-enhanced}. % for data-to-text generation. 
%More recently, 
\citet{kanthara2022chart} released an extended dataset crawled from the same platform, also called Chart-to-text (c2t-big in this paper). They finetune multiple pretrained models, such as BART \citep{lewis2019bart} and T5 \citep{raffel2019exploring}. 
% Another recent dataset is Barch \citep{krjanec-edhi-demberg:2022:LREC}. It is interesting in that each chart is associated with the underlying data and several human-written summaries per chart, however, it turned out to be too small for effectively training our model.
Table \ref{tab:Dist1} shows the statistics of both datasets. 
%The datasets we use in this work are Chart-to-Text (c2t-small) by \citet{obeid2020chart}, and Chart-to-text (c2t-big) by \citet{kanthara2022chart}. 
\begin{table}[t] % Apparently the asterisk makes it centered (instead of being in one column)
\small\centering
\begin{tabular}{lrrrr}  % Here you need to declare how many columns you want to have. The c means that the content is centered. There is also r and l for right and left alignment
\toprule
  \textbf{Dataset} & \textbf{Training} & \textbf{Validation} & \textbf{Test} & \textbf{Total} \\\midrule
  c2t-small & 5,703 & 1,222 & 1,222 & 8,147 \\
  c2t-big & 24,367 & 5,222  & 5,222 & 34,811 \\
  %Barch \citep{krjanec-edhi-demberg:2022:LREC} &  1,063 & 660 & 213 & 190 \\
\bottomrule
\end{tabular}
\caption{Dataset sizes and splits: c2t-small by \citet{obeid2020chart} and c2t-big by \citet{kanthara2022chart}.}
\label{tab:Dist1}
\end{table}

\subsection{Hallucinations in Data-to-text NLG}
\label{sec:halluc}

In NLG, hallucination or unfaithful output means generated text that is not grounded in the input. \citet{ji2022survey} describe two types of hallucinations: \textbf{Intrinsic Hallucinations} refer to generated output that contradicts the source content, and \textbf{Extrinsic Hallucinations} refer to output that cannot be verified by the source. \citet{ji2022survey} name three main causes for hallucinations: (1) source-reference divergence (reference text not supported by the input data), (2) modeling choices, and (3) decoding strategies.
Efforts are made to mitigate hallucinations in NLG, particularly in data-to-text models. At the data level, clean and faithful datasets such as ToTTo \citep{parikh2020totto} and RotoWire-FG \citep{wang2020revisiting} were created with significant manual effort. For data preprocessing, \citet{nie-etal-2019-simple} use a domain-specific language understanding module to automatically remove noise from input meaning representations and reduce hallucinations. 
A more domain-general automatic approach was proposed by \citet{duvsek2020evaluating} with a transformer finetuned for NLI, but it was only applied to evaluate faithfulness of generated text, not to correct it.
Our approach combines the latter two by applying NLI for automatic data cleaning.
%NLI has also been used for measuring faithfulness of generated texts. For data-to-text generation, \citet{duvsek2020evaluating} used a transformer finetuned for NLI to evaluate generated text. 
%The generated text is said to be true if it mentions all and only the input facts.

\section{Problem Identification}
\label{sec:problem}

%\subsection{Hallucinations in Generated Summaries}\label{prob-1}
We manually analyzed two samples of 50 generated chart summaries of \citet{obeid2020chart}'s transformer trained on c2t-small and \citet{kanthara2022chart}'s T5 model finetuned on c2t-big.
For the c2t-small transformer, 22 summaries had intrinsic and 13 had extrinsic hallucinations. Some summaries were also incoherent and repetitive. For the c2t-big T5 model, 4 out of 50 summaries had intrinsic and 11 had extrinsic hallucinations.

Based on our analysis, we speculate that intrinsic and extrinsic hallucinations have different causes as identified by \citet{ji2022survey}. Intrinsic hallucinations may happen due to input sequence formatting (i.e., modeling choices), while extrinsic hallucinations are caused by the source-reference divergence in training data (missing input information and noise) \citet{maynez-etal-2020-faithfulness}. We address the input format in Section~\ref{sec:linearization} by adding context and reducing long-distance dependencies, and we further address noise in references in Section~\ref{sec:nli-filt}.

\section{Input Format Adjustment}
\label{sec:linearization}

\subsection{Context and Distance in Input Formatting}
\label{exp-1}

The source data table must be linearized for input into a sequence-to-sequence model.
\citet{obeid2020chart}'s linearization includes $x$ and $y$ axis labels, values, and chart type, but lacks the chart title (see Table~\ref{tab:input_lin_summary-1} in the Appendix for an example). We speculate that excluding the title results in extrinsic hallucination, i.e., generation of entities from parametric knowledge instead of the input data \citep{longpre-etal-2021-entity}. 

\citet{kanthara2022chart}'s T5-based approach produced far better results than \citet{obeid2020chart}, but some hallucinations were still present. They format the data table by following the template: \textit{title + y-values + x-values} (see Table~\ref{tab:input_lin_summary-2} in the Appendix for an example). This format includes the title, but it lacks the $x$ and $y$ axis labels and the corresponding $x$-$y$ values are not adjacent. The distance between each $x$ and its corresponding $y$ value is large, and we speculate the model faces difficulty when learning pairwise relationships between $x$ and $y$, leading to intrinsic hallucination.

\begin{table}[t] % Apparently the asterisk makes it centered (instead of being in one column)
    \small\centering
    \begin{tabular}{|p{0.9\linewidth}|} % Here you need to declare how many columns you want to have. The c means that the content is centered. There is also r and l for right and left alignment
    \hline % This is how horizontal lines are added, you can put them anywhere  e
    Road rage behavior among drivers in the U.S. as of 2015 x-y labels situation - share of respondents x-y values On the receiving end of a rude gesture 53\%, Yelled or used profanity 26\%, Made a rude gesture 17\%, Felt physically threatened 13\%, Exited their vehicle to engage angrily 4\%\\		
    \hline
    \end{tabular}
    \caption{Example of our proposed linearization (See the chart in Table \ref{tab:intro-example}). After the chart title, the input contains $x$ and $y$ axis labels, followed by \textit{x-y} pairs, such that each $y$ value is adjacent to its corresponding $x$ value.}
    \label{tab:input_lin_example}
\end{table}

\subsection{Proposed Input Formatting}
\label{sec:proposed-format}
Considering these input format problems, we hypothesise that reducing long-distance dependencies between $x$ and $y$ axis values in the linearized input data will alleviate intrinsic hallucinations; adding title and $x$ and $y$ axis labels should reduce extrinsic ones.
We thus propose a linearized input with adjacent \textit{x-y} pairs. The template we use is: \textit{title + x-y labels + x-y values}. See Table \ref{tab:input_lin_example} for an example.

\subsection{Experimental Setup}
\label{sec:linearization-experiment}

We finetune T5 \cite{raffel2019exploring} with our linearization proposed in Section~\ref{sec:proposed-format}, 
comparing to both original linearizations discussed in Section~\ref{exp-1}.
%on two different datasets: \textit{T5-S-OL} on c2t-small and \textit{T5-B-OL} on c2t-big.
We include ablated versions to check the effects of including the title, including axis labels, or using adjacent \emph{x-y} pairs.
As a prefix to T5's decoder input, we use \textit{``C2T:~''}. More training details are provided in  Table~\ref{tab:hyperparameters} in the Appendix.
%before every instance of input data. 
We evaluate using BLEU \citep{post-2018-call}, ROUGE-2 \citep{lin2004rouge}, perplexity,\footnote{\url{https://huggingface.co/docs/transformers/perplexity}} and NUBIA \citep{kane-etal-2020-nubia}. NUBIA produces a score based on logical agreement, contradiction, neutrality, and semantic similarity. 

% We train three chart summarization models by finetuning T5 \cite{raffel2019exploring}: (1) \textit{T5-S-O\&HL} is finetuned on c2t-small linearized following the method by \citet{obeid2020chart}. % (see Table~\ref{tab:input_lin_summary-1}). 
% (2) \textit{T5-S-OL} (Our Linearization) was finetuned on c2t-small using our linearization proposed in Section~\ref{sec:proposed-format}.
% %as shown in Table~\ref{tab:input_temp}. 
% (3) \textit{T5-B-OL} was finetuned with our proposed linearization on the c2t-big data (see Table~\ref{tab:Dist1} for datasets). 
% %The splits for both the datasets are given in Table \ref{tab:Dist1}. 

\subsection{Metrics Results}
\label{sec:linearization-metrics}

\begin{table*}[t] % Apparently the asterisk makes it centered (instead of being in one column)
    \small\centering
    \begin{tabular}{lrrrrrrrr} % Here you need to declare how many columns you want to have. The c means that the content is centered. There is also r and l for right and left alignment
        \toprule
          \textbf{Model} & \textbf{BL}$\uparrow$ & \textbf{RG-2}$\uparrow$ &\textbf{PPL}$\downarrow$ & \textbf{L}$\uparrow$ & \textbf{C}$\downarrow$ & \textbf{Neu} & \textbf{SS}$\uparrow$ & \textbf{N}$\uparrow$ \\*\midrule
          Transformer by \citet{obeid2020chart} &  18.5 & - & - & - & - & - & - & - \\
          T5-S-O\&HL   & 26.1 & 33.5 & 7.4 & 5.5 & 67.8 & 26.5 & 3.0/5 & 35.4 \\
          T5-S-O\&HL + title & 31.0 & 44.0 & 16.0 & 24.8 & 23.5 & 51.5 & 3.0/5 & 60.6 \\
          T5-S-OL &  33.9 & 44.8 & 7.5 &33.2 & 22.3 & 44.4 & 3.5/5 & 46.9  \\
          T5-S-OL-NLI & 34.2 & 43.7 & 7.1 & 33.1 & 10.2 & 56.5 & 3.5/5 & 44.5 \\
        \bottomrule
        \end{tabular}
    \caption{Evaluation results on c2t-small for input format improvements (Section~\ref{sec:linearization}) and NLI filtering (Section~\ref{sec:nli-filt}): BLEU-4 (BL), ROUGE-2 (RG-2), Perplexity (PPL), Logical Agreement (L), Contradiction (C), Neu (Neutrality), Semantic Similarity (SS) and the NUBIA (N) score.}
    \label{tab:result-1}
\end{table*}

\begin{table*}[t] % Apparently the asterisk makes it centered (instead of being in one column)
    \small\centering
    \begin{tabular}{lrrrrrrrr} % Here you need to declare how many columns you want to have. The c means that the content is centered. There is also r and l for right and left alignment
    \toprule % This is how horizontal lines are added, you can put them anywhere
      \textbf{Model}&\textbf{BL}$\uparrow$ &\textbf{RG-2}$\uparrow$ & \textbf{PPL}$\downarrow$ & \textbf{L}$\uparrow$ & \textbf{C}$\downarrow$ & \textbf{Neu} & \textbf{SS}$\uparrow$ & \textbf{N}$\uparrow$ \\*\midrule
      T5 by \citet{kanthara2022chart} & 37.0 & 50.5 & 10.0 &  34.5 & 22.9 & 42.5 & 3.6/5 & 53.5 \\ % BLEURT 0.15 
      T5-B-K + axis labels & 37.6 & 50.5 & 8.2 & 33.0 & 23.9 & 42.9 & 3.6/5 & 51.4\\
      T5-B-OL & 39.8 & 55.0 & 8.2 & 39.3 & 21.3 & 39.3 & 3.6/5 & 55.6\\ % BLEURT 0.17 
      T5-B-OL-NLI & 42.2 & 50.7 & 8.2 & 40.3 & 15.1 & 44.5 & 3.6/5 & 53.5\\ % BLEURT 0.17
    \bottomrule
    \end{tabular}
    \caption{Evaluation results for comparing linearization methods on c2t-big (see Table \ref{tab:result-1} for metrics).}
    \label{tab:result-2}
\end{table*}

On the c2t-small data, \textit{T5-S-OL} (our linearization) is compared to the original model of \citet{obeid2020chart}, a T5 finetuned using their linearization (\textit{T5-S-O\&HL}), and an ablation variant which uses their linearization and adds the chart title (\textit{T5-S-O\&HL + title}).
Results in Table~\ref{tab:result-1} show that our linearization %reducing long-distance dependencies and adding more chart information in the linearized input 
improves 
%\ODdel{BLEU, ROUGE-2, perplexity, logical agreement, contradiction, and semantic similarity.}
almost all metrics. While the overall NUBIA score is lower, its most important elements (logical agreement, contradiction, semantic similarity) are improved (cf.~Section~\ref{sec:nubia}).
\citet{obeid2020chart}'s input format produces many entity hallucinations. 
%, hampering performance. 
Including the chart title format improves performance substantially, which is expected as this provides crucial context for the model. Further small gains stem from less redundancy in our linearization.\footnote{Axis labels and chart type are not repeated in our format, compare Table~\ref{tab:input_lin_summary-1} and~\ref{tab:input_temp} in the Appendix.}

On the c2t-big data, \textit{T5-B-OL} (our linearization) is compared to the original T5 model of \citet{kanthara2022chart} and an ablation using their linearization with added axis labels (\textit{T5-B-K + axis labels}).
Table~\ref{tab:result-2} shows improvements on 
%BLEU, ROUGE-2, logical agreement, and contradiction}
almost all metrics, with NUBIA not reflecting its individual elements' improvements, similar as above (cf.~Section~\ref{sec:nubia}).
Adding axis labels to \citet{kanthara2022chart}'s format is a very modest help, but using adjacent \emph{x-y} pairs in our format yields a larger improvement.\footnote{More on ablations in Appendix~\ref{sec:ablations}.}

\subsection{Manual Analysis}
\label{sec:linearization-analysis}

We manually analyzed 50 output samples from T5-S-OL, checking for hallucinations. 
To find intrinsic hallucinations, we checked for any information in the summaries that would conflict with the input (x-y values, entities, or trends).
For extrinsic hallucinations, we checked for the presence of any information that was not verifiable from the input data.
%The examination of intrinsic hallucinations involved the assessment of x-y values, entities, and trends in the summaries. Summaries were examined for extrinsic hallucinations by verifying whether any information not present or that cannot be inferred by the input data, was present in the summary.  
We found no intrinsic hallucinations, but 18 summaries still had extrinsic hallucinations. Table~\ref{tab:example-summaries} in the Appendix provides example outputs. % from T5-S-OL and T5-B-OL.} 
%While these generations are factually correct, they still contain ungrounded information. 

% \subsection{Ablations}
% \label{sec:ablations}

%We conducted ablation experiments to investigate where the improvement comes from. We computed two ablations:  
% (1) \citet{obeid2020chart}'s linearization with added chart title, and (2) \citet{kanthara2022chart}'s linearization with added axis labels. 
% Results are shown in Tables~\ref{tab:result-1} and~\ref{tab:result-2}.
% Both ablation show improvements over the original linearizations, but do not match our scores, showing that all our format changes are helpful.
%The results for ablations are shown in the Appendix \ref{sec:appendix}. (1) resulting scores are between our linearization and \citet{obeid2020chart}'s linearization, and (2) resulting scores are very close to the original results, however, they are slightly worse than ours. 
%The biggest improvement is from (1) which makes sense because the more context the model has, the better the generation results. Similarly, adding x-y labels results in improvement compared to original results, but pairing the input data values together results in even better performance. 
%Overall, each of our modification in the input linearization leads to better performance, the biggest of which is including title of the chart in the input which provides more context to the model during training and inference. 

\section{Cleaning Noisy References with NLI}
\label{sec:nli-filt}

\subsection{Noise in Training Summaries}\label{prob-2}
Since source-reference divergence can also cause hallucinations, we analyzed the reference side of the same sample of 50 instances from the c2t-small dataset as in Section~\ref{sec:problem} to look for text that is not grounded in the source chart.
20 out of 50 summaries contained ungrounded information. While this ungrounded information makes the summaries more interesting, it cannot be verified from the chart and hence counts as extrinsic hallucination. We also analyzed references in the c2t-big dataset and found a similar pattern, which is expected since both datasets come from the same source. 

\subsection{Influence on Generation}
\label{sec:autochart}

To show that ungrounded information in training data influence system outputs, we run an experiment on the Autochart dataset \cite{zhu2021autochart}, which is handcrafted and thus guaranteed not to contain hallucinations.
We introduce synthetic ungrounded text at random places in Autochart summaries using vanilla GPT-2 \cite{radford2019language} generation prompted by preceding summary text, thus creating a noisy Autochart version.\footnote{See Section \ref{subsec:appendix-1} in the Appendix for details.} The summary of the chart is segmented, and a random sentence from the summary is used as a prompt for GPT-2 to produce an ungrounded sentence. The generated text is then inserted at a random location in the segmented summary, creating a new summary with ungrounded information.
We analyze 50 outputs from a T5 model finetuned on both original (T5-AC-orig) and noisy (T5-AC-noisy) versions. %\textsuperscript{\ref{fn:hyperparams}} 
While we found no hallucinations in outputs of T5-AC-orig, we identified 27 extrinsic hallucinations in T5-AC-noisy's outputs. T5-AC-noisy also produced repetitive summaries (43 out of 50).

\subsection{Improving Faithfulness using NLI}\label{nli-explanation}

To alleviate extrinsic hallucinations caused by the training data, we propose using NLI, taking inspiration from \citet{pang2021agreesum} and \citet{duvsek2020evaluating}. We use NLI as a preprocessing tool: any sentences in a summary that are not entailed in the linearized data will be discarded. 
We use the BART-MNLI pretrained model\footnote{\url{https://huggingface.co/facebook/bart-large-mnli}} for this.
It is based on \citet{yin-etal-2019-benchmarking}'s study on zero-shot text classification as an entailment problem and trained to produce an entailment score on a scale of 0-100 (with no specific neutral or contradiction labels).
%\citet{yin-etal-2019-benchmarking} studied zero-shot text classification as an entailment problem, and based on this study, zero-shot BART-MNLI\footnote{\url{https://huggingface.co/facebook/bart-large-mnli}} was developed. We employ the pretrained BART NLI model for dataset cleaning. The model produces a confidence score between 0-100, where 0 indicates no entailment, and 100 indicates entailment.
%\OD{I rephrased this to make it a bit more concise, please check.}

\begin{figure}[t]
    \centering
    \includegraphics[width=0.8\linewidth]{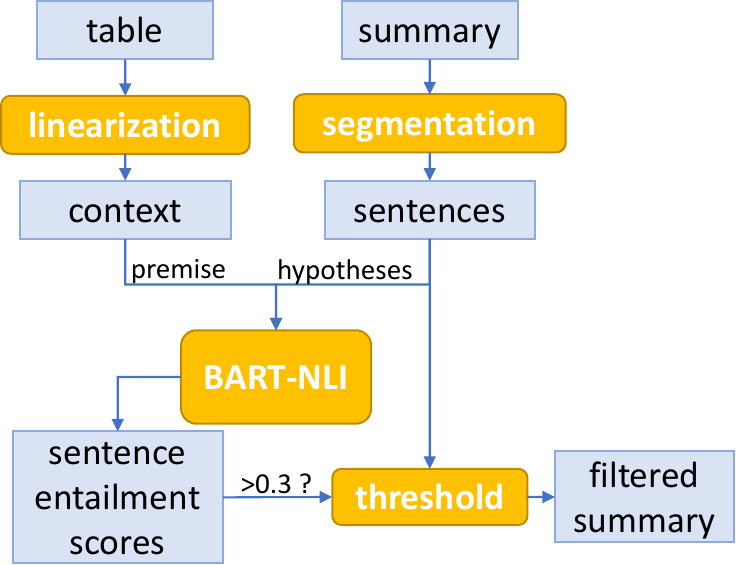}
    \caption{Summary cleaning process using zero-shot BART-NLI.}
    \label{fig:BART-NLI-filtering}
\end{figure}

Figure \ref{fig:BART-NLI-filtering} shows a diagram of the overall cleaning process. We apply the filtering step to the entire c2t-small dataset by segmenting the summaries, passing each sentence as hypothesis and linearized data as premise. If the sentence gets an entailment score above the threshold of 0.3, we keep it, otherwise we discard the sentence.\footnote{The threshold was found empirically (see Appendix \ref{subsec:appendix-2}).} 
Out of the 100 filtered summaries, 69 were correctly filtered, and 31 were incorrectly filtered. We also applied the above process to c2t-big dataset. 

\subsection{Finetuning T5 on Filtered Data}\label{fine-tune-T5}

To evaluate the filtering, we compare two pairs of finetuned T5 models on both c2t-small and c2t-big, filtered (T5-B/S-OL-NLI) and non-filtered (T5-B/S-OL). All use 
%: (1) trained on the filtered dataset (T5-S-OL-NLI) and the other on the non-filtered one, and (2) trained on the filtered dataset (T5-B-OL-NLI) and the other on the non-filtered one. We use the c2t-small dataset for (1), c2t-big dataset for (2), and
our proposed linearization method. We use data splits described in Section \ref{tab:Dist1}.
%\subsection{Results and Error Analysis}

Tables~\ref{tab:result-1} and~\ref{tab:result-2} show the results of T5 finetuned on the filtered data and the baselines. The automatic metrics do not show any clear trend for T5-S-OL-NLI: BLEU improves but ROUGE drops; NUBIA logical agreement stays unchanged, contradictions decrease but neutral statements (also interpretable as hallucinations) increase. For T5-B-OL-NLI, we get minor improvements in BLEU and NUBIA logical agreement, but again a drop in ROUGE.
%The improvement in contradiction scores for both models indicates that extrinsic hallucinations have been alleviated significantly.\OD{It doesn't really, you can't claim this -- most of the contradictions move to neutral, which is still a likely hallucination. I keep telling you this.} 
This does not indicate improvements; however, upon manual analysis of 50 summaries
%We also manually analyzed 50 summaries
for the same charts as in Section~\ref{sec:linearization-analysis},
we found that only 4 outputs from T5-S-OL-NLI still contained hallucinations (compared to 18 for T5-S-OL).

\begin{table}[t] % Apparently the asterisk makes it centered (instead of being in one column)
    \centering\tabcolsep=5pt
    \small
        \begin{tabular}{l>{\hspace{-2mm}}c>{\hspace{-1mm}}c>{\hspace{-1mm}}c>{\hspace{-1mm}}c>{\hspace{-1mm}}c>{\hspace{-1mm}}c}
        \toprule
          \textbf{Model} & \textbf{VC}$\uparrow$ & \textbf{OIP}$\downarrow$ & \textbf{Inf}$\uparrow$ & \textbf{Coh}$\uparrow$ & \textbf{Flu}$\uparrow$ \\ 
          \midrule
          T5-S-OL & 56.00\% & 38.00\% & 3.80/5 & 3.81/5 & 3.88/5\\*
          T5-S-OL-NLI & 76.00\% & 17.00\% & 3.60/5 & 3.91/5 & 3.96/5\\*
        
            %$p$ & 0.000015 & 0.00001 & 0.9136 & 0.333218 & 0.4648 \\
            $p$-value & 1.5e-5 & 1.0e-5 & 0.914 & 0.333 & 0.465 \\
        \bottomrule
        \end{tabular}
    \caption{Human evaluation results (see Section~\ref{sec:human-eval}). We used a $\chi^2$ test for VC and OIP, and one-way ANOVA for Inf, Coh, and Flu.}
    \label{tab:human-eval}
\end{table}

\subsection{Human Evaluation}
\label{sec:human-eval}

We conduct a detailed human evaluation, comparing T5-S-OL (see Section~\ref{exp-1}) and T5-S-OL-NLI (trained in Section~\ref{fine-tune-T5}). We evaluate the following: 
(1) \textbf{Value Correctness (VC)}: Numbers/values in the summary are from the chart, %The annotator determines which of the summaries are accurate, 
(2) \textbf{Outside Information Presence (OIP)}: The summary contains information not grounded in the chart,  (3) \textbf{Informativeness (Inf)}: The summary conveys a lot of information about the chart, (4) \textbf{Coherence (Coh)}: The summary content is orderly and logically consistent, and (5) \textbf{Fluency (Flu)}: The text is grammatically correct and is not repetitive.

We used Prolific to recruit English native speakers from the UK\footnote{\url{https://www.prolific.co/}, the hourly pay rate was 9.5~GBP.} and Google Forms to conduct the survey. For each model, 50 samples were used and split into 5 experiments with 10 samples each. Each sample was annotated by 5 participants, in total 25 participants completed the survey. Table \ref{tab:human-eval} shows the result. Considering faithfulness (VC and OIP), the T5-S-OL-NLI model trained on filtered data performs significantly better than the baseline T5-S-OL, showing that our method of alleviating hallucinations via cleaning training summaries through NLI
%and adding more context in the linearization 
is effective.
%\OD{You can't claim this for linearization changes, you're not comparing to previous linearizations here!} 

\section{Discussions}

%\OD{Can you point me where this came from? Is this a reaction to reviewers' comments?}
%\OD{I believe both sections should be shortened a bit; 6.2 is kinda reflected in the text already}

\subsection{Ungrounded Information in Training Data}
\label{sec:ungrounded}

In Section~\ref{prob-2}, we reported on ungrounded information in training data and showed in Section~\ref{sec:autochart} that this leads to hallucinations in generated outputs. 
While there are good reasons for ungrounded information in human-written summaries (e.g., providing additional detail/background or linking to other events; cf.~\citealp{thomson_gold_2020}), using such data to train an end-to-end model that does not distinguish between describing the chart and providing additional information is not appropriate and leads to inaccurate outputs, which is unsuitable for real-world scenarios \cite[cf.~][]{maynez-etal-2020-faithfulness,xu_miranews_2021}.

\subsection{Metrics' Shortcomings in Assessing Hallucinations}
\label{sec:nubia}

None of the automatic metrics we used (see~Section~\ref{sec:linearization-experiment}) measure hallucinations explicitly.
BLEU and ROUGE are reference-based and prone to biases stemming from ungrounded information in references (see Section~\ref{sec:ungrounded}),
such as assigning higher scores to hallucinated outputs.
%The metrics we employed in our study do not measure hallucinations explicitly. BLEU and ROUGE are n-gram based metrics and do not measure the quality of texts, let alone hallucinations. 
NUBIA is a trained aggregate metric of several components (NLI-based logical agreement, contradiction and neutrality, plus semantic similarity and fluency). We found that this aggregation can lead to nontransparent or confusing results: even if individual components are clearly improved, NUBIA may drop (see Section~\ref{sec:linearization-metrics}).
%However, the final aggregation function (a neural network) that computes the NUBIA (N) score results in intransparent results. 
%Sometimes, a clear pattern emerges that indicates there is an improvement (e.g.~comparing T5-S-O\&HL and T5-S-OL) but other times, there is no clear pattern for improvements (e.g.~comparing T5-S-O\&HL, T5-S-OL and T5-S-OL-NLI).
Therefore, we found logical agreement %, contradiction 
and semantic similarity scores to be the most useful constituents of NUBIA for indicating the presence of hallucinations in generated texts. 
% This argument is the same essentially as what you said in the last paragraph of 6.1
%Aside from the interpretation difficulties related to neural-network-based NLG metrics, another short-coming is that all of them are reference-based metrics. The problem with reference based metrics is the reference itself, which contains ungrounded information, which may lead to evaluation bias: higher evaluation scores for outputs that align with flawed reference text. To this end, the research community should put in work to develop reference-free metrics for closed-ended NLG tasks.

\section{Conclusions}
We show that reducing long-distance dependencies and providing more context on the model's input results in fewer intrinsic hallucinations, and demonstrate that extrinsic hallucinations are a result of ungrounded information in the training summaries. Furthermore, we show through human evaluation that employing NLI to filter training summaries results in a significant drop in hallucinations.

\section*{Limitations}

% XXX would count toward page limit for INLG, TODO bring back in camera-ready
The main limitation of our work is that we were unable to eliminate the extrinsic hallucinations completely. In Section \ref{nli-explanation}, we mentioned that 31 out of 100 summaries were not filtered correctly, meaning that these summaries were left with ungrounded information, which resulted in 4 out of 50 generated summaries with extrinsic hallucinations. BART-NLI is developed for linguistic input and we employ it to infer from non-linguistic input. The second limitation is that generated summaries are shorter on average. T5-S-OL-NLI generated 28\% of the summaries of just a single sentence. We expected this problem because our method of filtering only removes sentences and does not replace them with statements entailed in the data. The final limitation is that our model is only limited to producing summaries in the English language as it is trained on English summaries.

\section*{Acknowledgements}

This research was supported by the European Research Council (Grant agreement No. 101039303 NG-NLG) and (in part) supported by the EUINACTION grant from NORFACE Governance (462-19-010, GL950/2-1).  
We acknowledge the use of resources provided by the LINDAT/CLARIAH-CZ Research Infrastructure (Czech Ministry of Education, Youth and Sports project No.~LM2018101).

\section*{Ethics Statement}
The human evaluation study was approved by the ethics committee of the respective national professional linguistic association.
All the annotators were from the United Kingdom and each annotator was paid according to the hourly minimum wage in the United Kingdom, i.e.\ 9.5~GBP. The annotators were paid immediately after the results were analyzed. We only collected Prolific IDs of the users and they were deleted after the analysis of the data.

\bibliography{acl2023}
\bibliographystyle{acl_natbib}

\appendix
\section{Appendix}\label{sec:appendix}

\subsection{Experiment using Autochart}\label{subsec:appendix-1}
\subsubsection{Splits}
The authors of autochart did not create any data splits. The total size of the dataset is 23,543 chart-summary pairs. From the dataset, we use 10,593 and split it with the ratio of 70:15:15. 

\subsubsection{GPT-2 Noise Generation}
To inject noise in the summaries, we first segment the summary using NLTK \citep{bird2009natural} sentence tokenizer. After segmenting the summary, we randomly pick a sentence and give it as a prompt to the GPT-2 model. For GPT-2 generation, we use greedy search. The generated sentence is then inserted at a random location in the segmented summary list, and then all the sentences are combined.

\subsubsection{Threshold Determination}\label{subsec:appendix-2}
We analyzed a random sample of 100 filtered summaries and found that the mean entailment score of the entailed sentences was 89, while the mean entailment score given to the non-entailed sentences was 8.7. This means that the model is sure when assigning the score, and making minor adjustments to the threshold would not lead to significant improvements.

\subsection{Ablations}\label{sec:ablations}

We conducted ablation experiments to investigate where the improvement comes from in the linearization. We computed two ablations:  
(1) \citet{obeid2020chart}'s linearization with added chart title (T5-S-O\&HL+title), and (2) \citet{kanthara2022chart}'s linearization with added axis labels (T5-B-K + axis labels). 
Results are shown in Tables~\ref{tab:result-1} and~\ref{tab:result-2}.
Both ablation show improvements over the original linearizations, but do not match our scores, showing that all our format changes are helpful. For (1) resulting scores are between our linearization and \citet{obeid2020chart}'s linearization, and for (2) resulting scores are very close to the original linearization results, however, they are slightly worse than ours. 
The most significant enhancement is due to (1), which is understandable as the model's performance improves with increased context. Likewise, the inclusion of x-y labels leads to an enhancement over the initial outcomes. However, the performance is further boosted when the input data values are combined.

\subsection{Human Evaluation Survey Details}\label{subsec:appendix-3}

\subsubsection{Consent Form}
Each user was asked to sign the consent form based on the following text:
This study is being conducted as part of ongoing research at [------]. If you have any questions or comments about the study, please contact us on Prolific. You must be at least 18 years old to participate. Your participation in this research is voluntary. There are no risks or benefits to participating in this study.  In the next section we will ask for your Prolific ID. All data will be anonymized prior to analysis and Prolific IDs will not be published.

\subsubsection{Survey Description} 
Dear Participants, you will be evaluating summaries of charts. Choose the summary that has Value Correctness and Outside Information Presence. Rate the informativeness, coherence, and fluency of the summaries given the chart.

\paragraph{Value Correctness:} Numbers/figures/values in the summary are from the chart. Here you determine which of the summaries are accurate. 
\paragraph{Outside Information:} Information that is not from the chart at all. Here you determine which of the summaries have information not taken from the chart. 
\paragraph{Informativeness:} The summary conveys a lot of information about the chart. 1 being the least informative and 5 being the most informative. 
\paragraph{Coherence:} The information included in the summary is orderly and logically consistent. Here you rate the coherence of the summary. 1 being the least coherent and 5 being the most coherent. 
\paragraph{Fluency:} Summary is grammatically correct and does not contain any repetitions. Here you rate the fluency of the summary. 1 being the least fluent and 5 being the most fluent.

\subsubsection{Evaluation of Measured Properties}
Value correctness gives us a binary scores, meaning, either the summary has correct values or not. Similarly for outside information presence, we also get binary scores. For, informativeness, coherence, and fluency, we get scores out of 5-point Likert scale \citep{likert1932}, 5 being the highest score, and 1 being the lowest score.

\begin{table*}[h!] % Apparently the asterisk makes it centered (instead of being in one column)
    \centering
    \begin{tabular}{p{0.28\linewidth}|p{0.35\linewidth}|p{0.28\linewidth}} % Here you need to declare how many columns you want to have. The c means that the content is centered. There is also r and l for right and left alignment
    % This is how horizontal lines are added, you can put them anywhere  
    \hfil\textbf{T5-S-OL} & \hfil\textbf{T5-B-OL} & \hfil\textbf{T5-S-OL-NLI}\\
    \hline
    This statistic shows road rage behavior among drivers in the United States. During the survey, 53 percent of respondents stated they had been on the receiving end of a rude gesture. \textit{All the participants in this survey had a valid U.S. driving license.} 
    &
    This statistic represents the road rage behavior among drivers in the United States as of April 2015. During the survey, 13 percent of respondents stated that they felt physically threatened by another driver to engage in angrily with another driver. \textit{The survey was conducted online and all the participants had a valid U.S. driving license.} &
    This statistic shows the road rage behavior among drivers in the United States. 53 percent of respondents said they had been on the receiving end of a rude gesture and 26 percent of the respondents said they have yelled or used profanity at another driver. 
    \\
    \end{tabular}
    \caption{Generated summaries from three different models for the chart in Table \ref{tab:intro-example}. The summaries from T5-S-OL and T5-B-OL contain \textit{extrinsic hallucinations}.} 
    \label{tab:example-summaries}
\end{table*}

\begin{table*}[h!] % Apparently the asterisk makes it centered (instead of being in one column)
    \centering\small
    \begin{tabular}{|p{0.5\linewidth}|p{0.5\linewidth}|} % Here you need to declare how many columns you want to have. The c means that the content is centered. There is also r and l for right and left alignment
    \hline % This is how horizontal lines are added, you can put them anywhere  e
      \raisebox{-4.5cm}{\includegraphics[width=\linewidth]{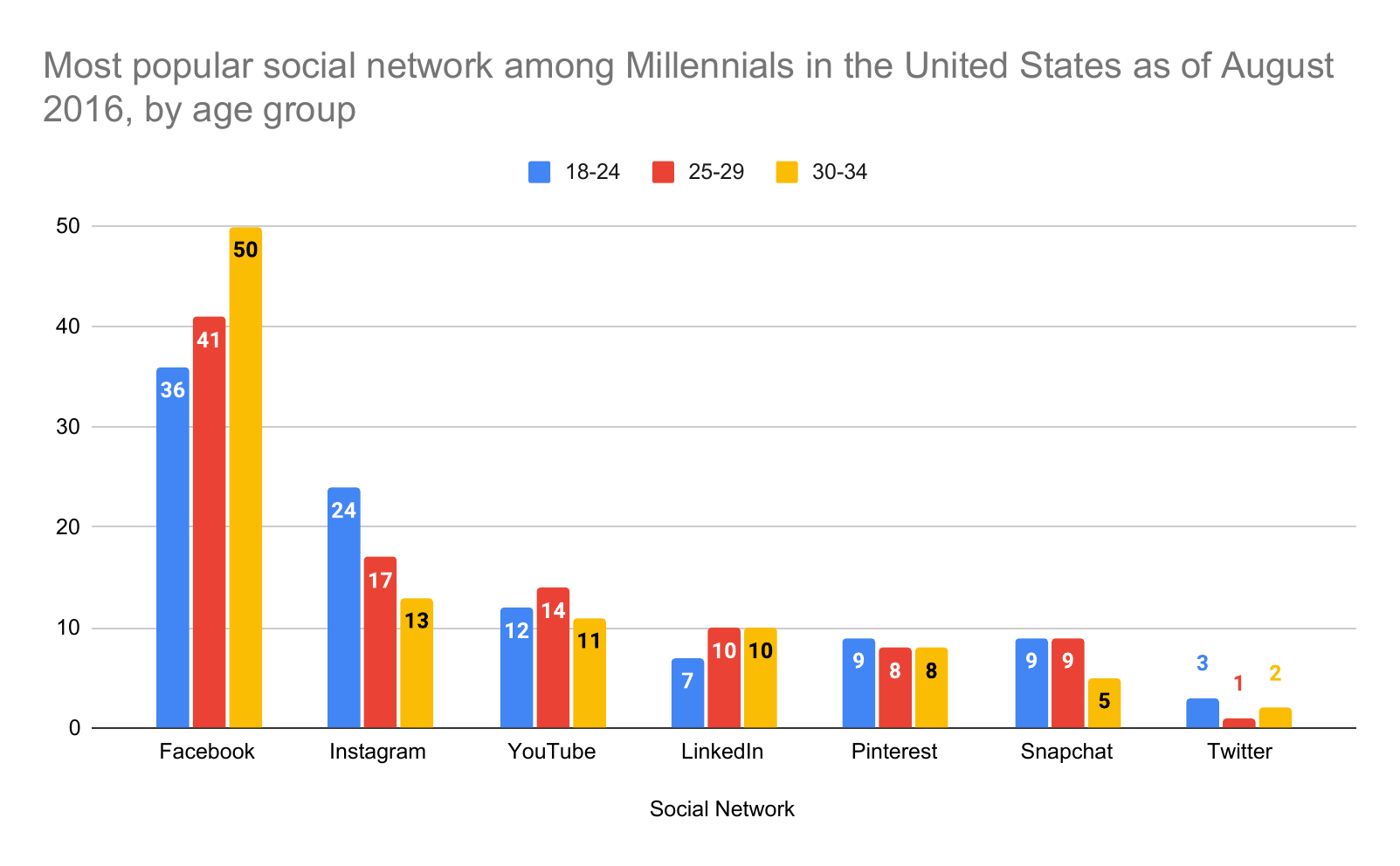}} & Platform $|$ Facebook $|$ 0 $|$ bar chart 18-24 $|$ 36 $|$ 1 $|$ bar chart 25-29 $|$ 41 $|$ 2 $|$ bar chart 30-34 $|$ 50 $|$ 3 $|$ bar chart Platform $|$ Instagram $|$ 0 $|$ bar chart 18-24 $|$ 24 $|$ 1 $|$ bar chart 25-29 $|$ 17 $|$ 2 $|$ bar chart 30-34 $|$ 13 $|$ 3 $|$ bar chart Platform $|$ YouTube $|$ 0 $|$ bar chart 18-24 $|$ 12 $|$ 1 $|$ bar chart 25-29 $|$ 14 $|$ 2 $|$ bar chart 30-34 $|$ 11 $|$ 3 $|$ bar chart Platform $|$ LinkedIn $|$ 0 $|$ bar chart 18-24 $|$ 7 $|$ 1 $|$ bar chart 25-29 $|$ 10 $|$ 2 $|$ bar chart 30-34 $|$ 10$|$ 3 $|$ bar chart Platform $|$ Pinterest $|$ 0 $|$ bar chart 18-24 $|$ 9 $|$ 1 $|$ bar chart 25-29 $|$ 8 $|$ 2 $|$ bar chart 30-34 $|$ 8 $|$ 3 $|$ bar chart Platform $|$ Snapchat $|$ 0 $|$ bar chart 18-24 $|$ 9 $|$ 1$|$ bar chart 25-29 $|$ 9 $|$ 2 $|$ bar chart 30-34 $|$ 5 $|$ 3 $|$ bar chart Platform $|$ Twitter $|$ 0 $|$ bar chart 18-24 $|$ 3 $|$ 1 $|$ bar chart 25-29 $|$ 1 $|$ 2 $|$ bar chart 30-34 $|$ 2 $|$ 3 $|$ bar chart   \\
    \hline
    \end{tabular}
    \caption{Linearized input format used by \citet{obeid2020chart}. Example from c2t-small dataset.}
    \label{tab:input_lin_summary-1}
\end{table*}

\begin{table*}[h!] % Apparently the asterisk makes it centered (instead of being in one column)
    \centering\small
    \begin{tabular}{|p{0.5\linewidth}|p{0.5\linewidth}|} % Here you need to declare how many columns you want to have. The c means that the content is centered. There is also r and l for right and left alignment
    \hline % This is how horizontal lines are added, you can put them anywhere  e
        \raisebox{-4cm}{\includegraphics[width=\linewidth]{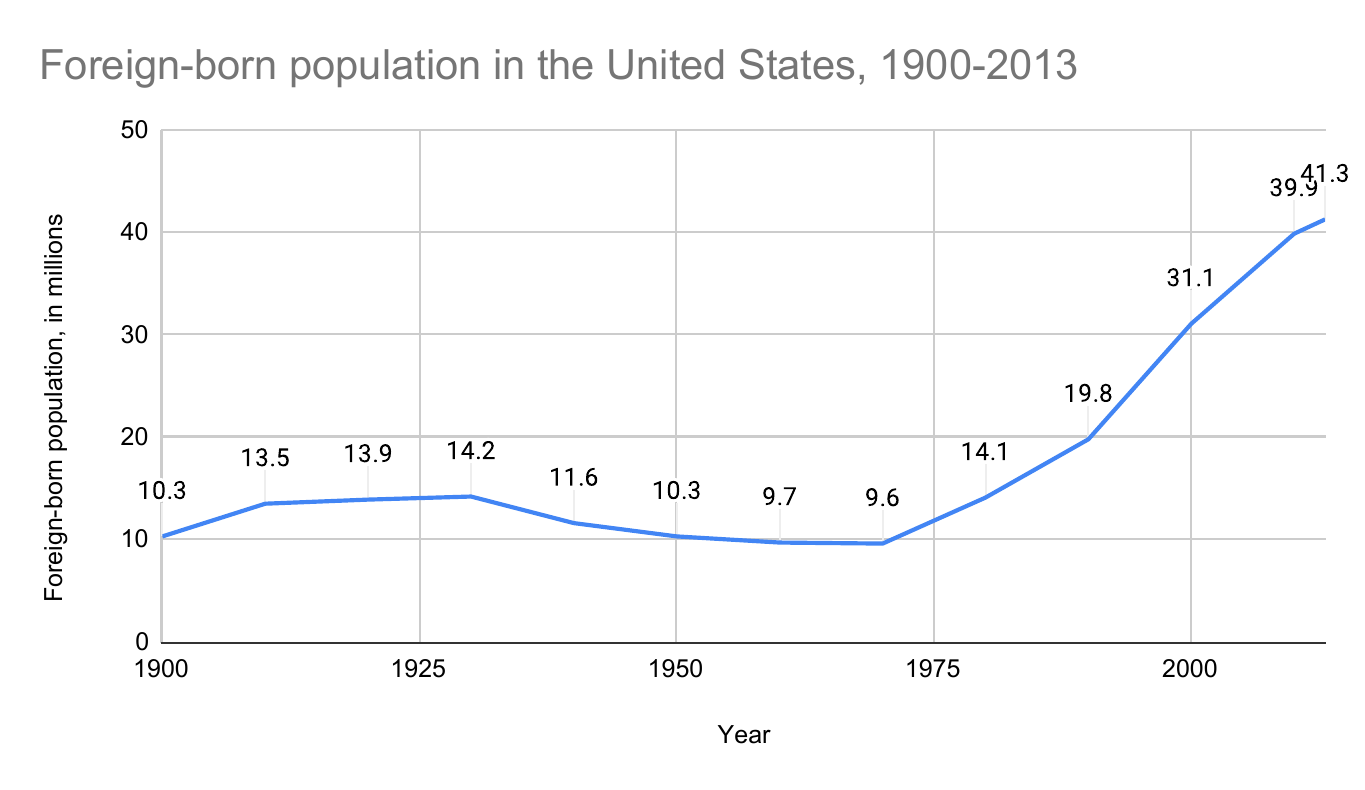}} & Foreign-born population in the United States, 1900-2013  10.3 $|$ 13.5 $|$ 13.9 $|$ 14.2 $|$ 11.6 $|$ 10.3 $|$ 9.7 $|$ 9.6 $|$ 14.1 $|$ 19.8 $|$ 31.1 $|$ 39.9 $|$ 41.3  $|$ 1900 $|$ 1910 $|$ 1920 $|$ 1930 $|$1940 $|$ 1950 $|$ 1960 $|$ 1970 $|$ 1980 $|$ 1990 $|$ 2000 $|$ 2010 $|$ 2013\\
    \hline
    \end{tabular}
    \caption{Linearized input format used by \citet{kanthara2022chart}. Example from c2t-big dataset.}
    \label{tab:input_lin_summary-2}
\end{table*}

\begin{table*}[h!] % Apparently the asterisk makes it centered (instead of being in one column)
    \centering\small
    \begin{tabular}{|p{0.6\linewidth}|p{0.4\linewidth}|} % Here you need to declare how many columns you want to have. The c means that the content is centered. There is also r and l for right and left alignment
    \hline % This is how horizontal lines are added, you can put them anywhere  
      \raisebox{-5cm}{\includegraphics[width=\linewidth]{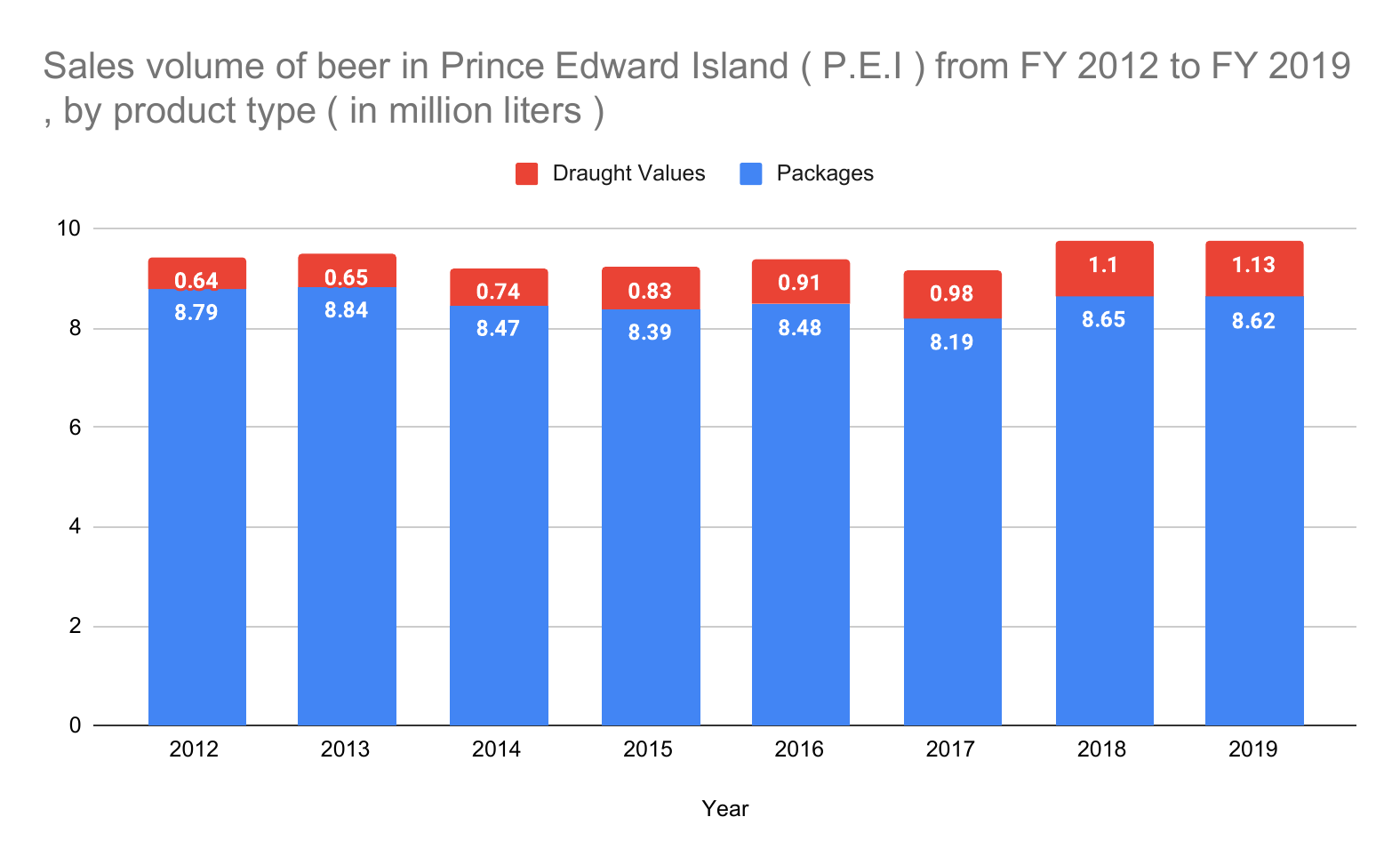}} & Sales volume of beer in Prince Edward Island ( P.E.I ) from FY 2012 to FY 2019 , by product type ( in million liters ) labels Year - Packaged - Draught values 2019 8.62 1.13 , 2018 8.65 1.1 , 2017 8.19 0.98 , 2016 8.48 0.91 , 2015 8.39 0.83 , 2014 8.47 0.74 , 2013 8.84 0.65 , 2012 8.79 0.64\\
    \hline
    \end{tabular}
    \caption{The proposed input format that we use for both the c2t-small and c2t-big dataset following the formula $title$ + $x$-$y$ labels + $x$-$y$ values. Individual parts are separated by a comma.}
    \label{tab:input_temp}
\end{table*}

\begin{table*}[t] % Apparently the asterisk makes it centered (instead of being in one column)
    \centering
    \begin{tabular}{|p{0.5\linewidth}|p{0.4\linewidth}|} % Here you need to declare how many columns you want to have. The c means that the content is centered. There is also r and l for right and left alignment
    \hline % This is how horizontal lines are added, you can put them anywhere  
      \textbf{Model Version} & \textbf{Model Repository} \\\hline  
       Pre-trained T5-base \citep{raffel2019exploring} & \url{https://huggingface.co/t5-base}\\
      \hline       
      \textbf{Parameter} & \textbf{Value} \\
      \hline
      Maximum input length   & 1024 \\
      \hline
      Maximum target length &  512 \\
      \hline
      Truncation & True  \\
      \hline
      Padding & max\_length  \\
      \hline
      batch size & 2  \\
      \hline
      Optimizer & Weighted Adam \citep{kingma2014adam} \\
      \hline
      Learning rate &  3e-4 \\
      \hline
      Weight decay & 0.01  \\
      \hline
      Training epochs and hours for T5-S-O\&HL & 6 epochs, 11 hours \\
      \hline 
      Training epochs and hours for T5-S-OL & 6 epochs, 11 hours  \\
      \hline
      Training epochs and hours for T5-AC-orig & 8 epochs, 6 hours \\
      \hline
      Training epochs and hours for T5-AC-noisy & 8 epochs, 6 hours \\
      \hline
      Training epochs for T5-S-OL-NLI & 6 epochs, 11 hours\\
      \hline
      Training epochs for T5-B-OL & 12 epochs, 37 hours \\
      \hline
      Training runs for all the models & Single run \\
      \hline
      Beam size & 4\\
      \hline
      GPU & Tesla T4 16 GB\\
    \hline
    \end{tabular}
    \caption{Hyper-parameters used and training details of our experiments.}
    \label{tab:hyperparameters}
\end{table*}

\end{document}